\documentclass{ifacconf}

\usepackage{graphicx}      
\usepackage{natbib}        

\usepackage{comment}
\usepackage{bm}
\usepackage{amsfonts}
\usepackage{amsmath}
\usepackage{amssymb}

\begin{document}
\begin{frontmatter}

\title{Distributed Shape Learning of Complex Objects Using Gaussian Kernel \thanksref{footnoteinfo}} 

\thanks[footnoteinfo]{This work was supported in part by Japan Society for the Promotion of Science (JSPS) KAKENHI under Grant 21K04104.}

\author[First]{Toshiyuki Oshima} 
\author[Second]{Junya Yamauchi} 
\author[Third]{Tatsuya Ibuki} 
\author[Fourth]{Michio Seto} 
\author[First]{Takeshi Hatanaka} 

\address[First]{School of Engineering, Tokyo Institute of Technology, 
   Tokyo 152-8552, Japan (e-mail: oshima.t@hfg.sc.e.titech.ac.jp; hatanaka@sc.e.titech.ac.jp).}
\address[Second]{Graduate School of Information Science and Technology
, The University of Tokyo,
   Tokyo 113-8656, Japan (e-mail: junya\_yamauchi@ipc.i.u-tokyo.ac.jp).}
\address[Third]{Department of Electronics and Bioinformatics, Meiji University, 
   Tokyo 214-8571, Japan (e-mail: ibuki@meiji.ac.jp).}
\address[Fourth]{Department of General Education, National Defence Academy, 
   Kanagawa 239-8686, Japan (e-mail: mseto@nda.ac.jp).}

\begin{abstract}
This paper addresses distributed learning of a complex object for multiple networked robots based on distributed optimization and kernel-based support vector machine.
In order to overcome a fundamental limitation of polynomial kernels assumed in our antecessor, we employ Gaussian kernel as a kernel function for classification.
The Gaussian kernel prohibits the robots to share the function through a finite number of equality constraints due to its infinite dimensionality of the function space.
We thus reformulate the optimization problem assuming that the target function space is identified with the space spanned by the bases associated with not the data but a finite number of grid points.
The above relaxation is shown to allow the robots to share the function by a finite number of equality constraints. 
We finally demonstrate the present approach through numerical simulations.
\end{abstract}


\end{frontmatter}

\section{Introduction}
Dynamic interactions between networked robots and objects in a mission space have been studied in the literature, including cooperative payload manipulation \citep{Spletzer:01, Dongjun:05} and obstacle avoidance \citep{Wang:17, Shibahara:22}.
While many studies have presented distributed algorithms to accomplish these missions, most of them assume prior knowledge of the exact shape, location, and other information about the target objects to be manipulated/avoided. This assumption may prevent the practical use of existing control methods due to the lack of such prior information.

In \citep{Oshima:22}, the authors addressed the above problem, where we presented a distributed shape learning method from point cloud data acquired by LiDAR (Light Detection And Ranging) sensors, based on the continuous-time alternating direction method of multipliers (ADMM) and the kernel-based support vector machine.
We also certified a data-independent performance for the ADMM algorithm through loop shaping techniques and passivity. 
This method however relies on a restrictive assumption that the kernel function is a polynomial kernel. 
Specifically, since the dimension of the function space for a given order is finite, a finite number of global equality constraints allow the robots to share the object shape.
Meanwhile, the polynomial function inherently suffers from a trade off that a low-order function may not fit a complex object shape while a high-order one increases communication/computation complexity and the risk of over-fitting.
Selecting an appropriate order \textit{a priori} contradicts the assumption that any prior knowledge of the object shape is not available.

In this paper, we employ a Gaussian kernel to avoid the limitations of the polynomial kernels. 
While the Gaussian kernel does not suffer from the above trade-off, it poses a new challenge.
Namely, the infinite dimension of the function space prohibits the robots to share the function through equality constraints.
To address the issue, we reformulate the optimization problem assuming that the target function space is identified with the space spanned by the bases associated with not the point cloud data but a finite number of grid points.
The above relaxation is shown to allow the robots to share the function by a finite number of equality constraints, if they share the grid points in advance. 
Finally, we verify the present approach through numerical simulations.

\section{Problem Formulation}

\subsection{Object Shape Learning by Multiple Robots}

\begin{figure}
    \begin{center}
    \includegraphics[width=8.4cm]{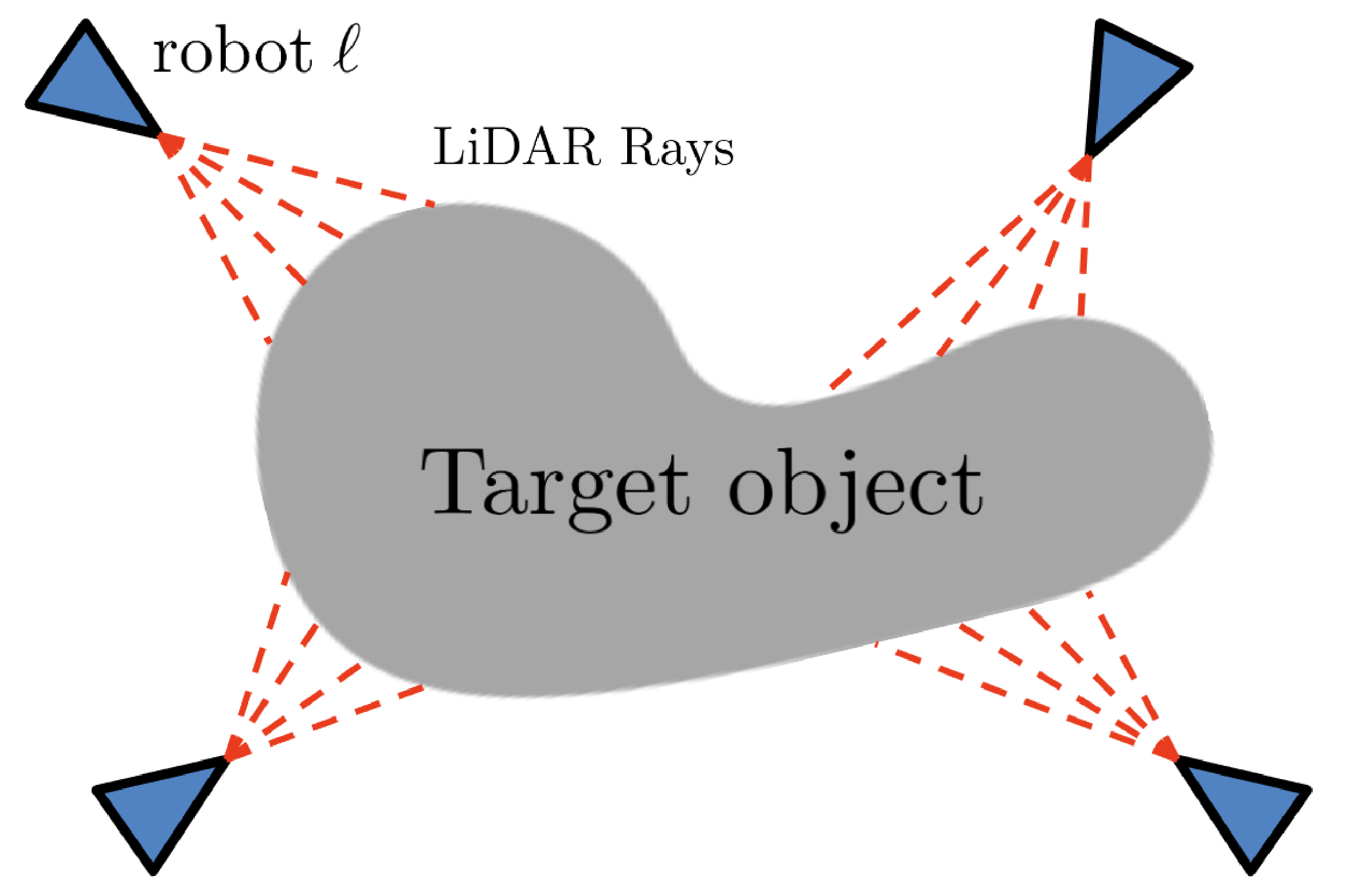}    
    \caption{Object shape learning by multiple robots.}
    \label{fig:scenario1}
    \end{center}
\end{figure}

Suppose that $N$ ground robots and an object whose shape must be learned are located on a 2-D plane $\mathcal{X} \subset \mathbb{R}^2$ as shown in Fig. \ref{fig:scenario1}.
Each robot has a LiDAR sensor that provides a finite number of points on the surface of the object. Since the LiDAR sensor has a limited sensing range, the point cloud acquired by each robot may concentrate on a part of the surface close to the robot.

\begin{figure}
    \begin{center}
    \includegraphics[width=8.4cm]{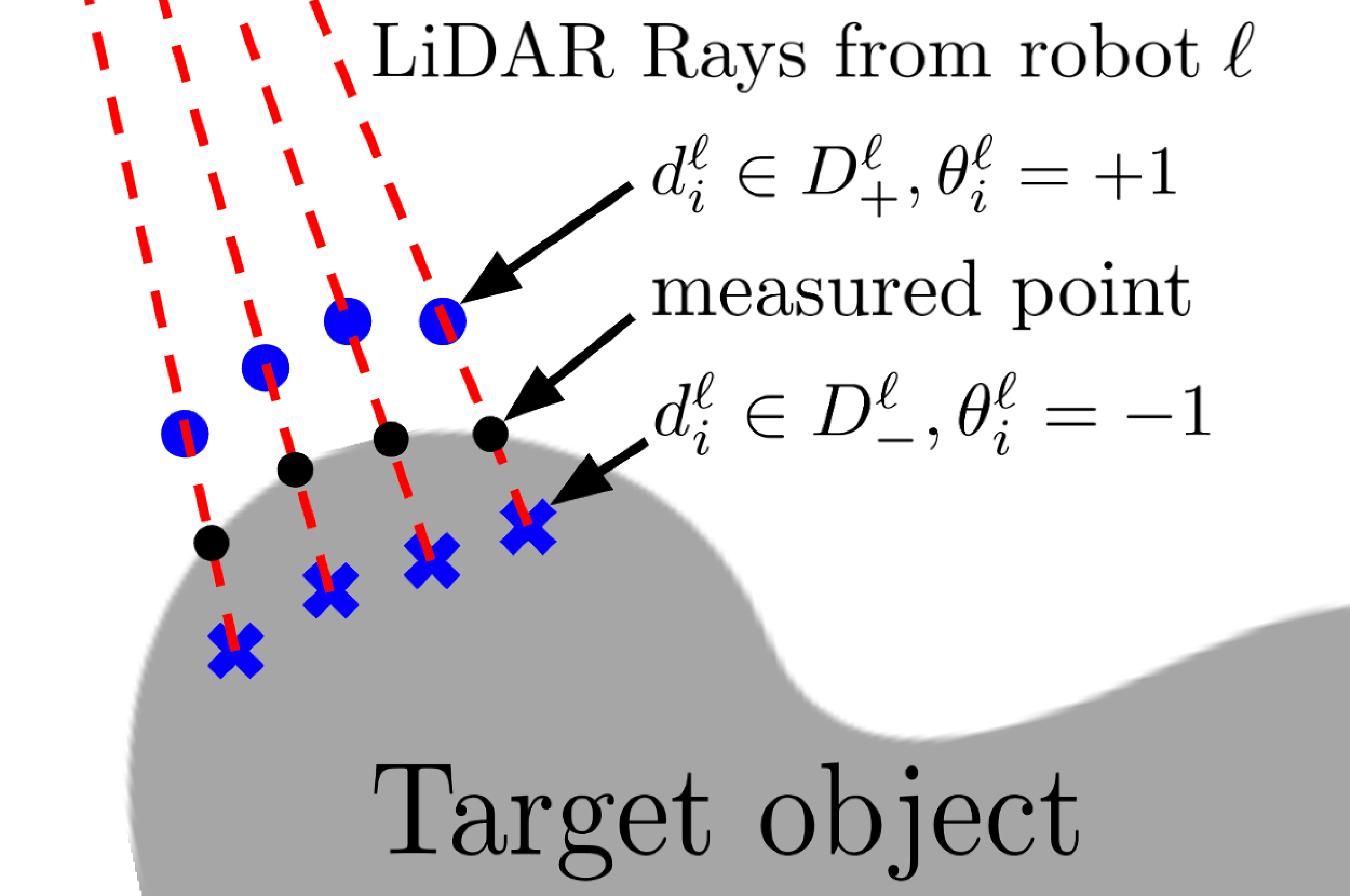}    
    \caption{Acquisition of data points by LiDAR sensor.}
    \label{fig:scenario2}
    \end{center}
\end{figure}

Now, suppose that robot $\ell\ (\ell = 1, 2, \ldots, N)$ generates $n^{\ell}$ points inside/outside of the object surface, as shown in Fig. \ref{fig:scenario2}.
Denote the positions of the data points as $d_{i}^{\ell} \in \mathcal{X}\ (i = 1, 2, \ldots, n^{\ell})$ in the world frame. The dataset acquired by robot $\ell$ is defined as $D^{\ell} := \{d_{1}^{\ell}, d_{2}^{\ell}, \ldots ,d_{n^{\ell}}^{\ell}\} \subset \mathcal{X}$.
Each data point $d_{i}^{\ell}$ has a unique label $\theta_{i}^{\ell} \in \{-1, +1\}$ such that $\theta_{i}^{\ell} = +1$ if $d_{i}^{\ell}$ is outside of the object, and $\theta_{i}^{\ell} = -1$ otherwise.
Accordingly, $D^{\ell}$ is divide into 
\begin{align*}
	D_{+}^{\ell} := \left\{d_{i}^{\ell} \in D^{\ell} \mid  \theta_{i}^{\ell} = +1 \right\}, \ D_{-}^{\ell} := \left\{ d_{i}^{\ell} \in D^{\ell} \mid  \theta_{i}^{\ell} = -1 \right\}. 
\end{align*}
Collecting all data points, define $D := \bigcup_{\ell = 1}^{N} D^{\ell}$, $D_{+} := \bigcup_{\ell = 1}^{N} D_{+}^{\ell}$, and $D_{-} := \bigcup_{\ell = 1}^{N} D_{-}^{\ell}$. 
The curve separating $D_{+}$ and $D_{-}$ is then expected to approximate the object surface. Therefore, the goal of this paper is to find such a separating curve, which corresponds to solving the following classification problem.
\begin{prob} \label{prob:class}
Find a function $f\in \mathcal{H}_k$ and $\gamma \in \mathbb{R}$ such that
\begin{align*}
	&D_{+} \subset \left\{ x \in \mathcal{X} \mid  f(x) + \gamma > 0  \right\}, \\
	&D_{-}  \subset \left\{ x \in \mathcal{X} \mid  f(x) + \gamma < 0  \right\},
\end{align*}
where $\mathcal{H}_k$ is the reproducing kernel Hilbert space (RKHS) \citep{Seto:22,Burges:98} associated with a kernel function $k(\cdot,\cdot)$.
\end{prob}
It is well-known that the problem is reduced to a quadratic program (QP) \citep{Seto:22}.

In our antecessor \citep{Oshima:22}, we reduced the QP into a form of distributed convex optimization assuming a polynomial kernel $k$, where the coefficients of the polynomial function $f$ are shared by the robots in the form of linear equality constraints.
We further presented a distributed optimization algorithm with data independent performance certificates by redesigning a continuous-time ADMM based on the concepts of passivity and loop shaping.
However, it is unclear if a polynomial function with a given order can fit a complex object shape.
In other words, the QP may be infeasible depending on the prescribed order of the polynomial function.

A promising solution to the above problem is using the Gaussian kernel
\begin{align}
    k(x,y) = \exp\{-\|x-y\|^2/2\}.
    \label{eqn:hatanaka_edit_gauss}
\end{align}
However, differently from the polynomial kernel, the space $\mathcal{H}_k$ cannot be characterized by a finite number of coefficients.
Accordingly, it is hard for the robots to share the shape with the other robots through a finite number of global equality constraints.
This is the issue addressed in this paper.

\subsection{Local Classification}

Problem \ref{prob:class} requires to collect of all data $D$ at a computer, which is not always realistic due to the large volume of the data provided by the LiDAR sensors. 
We thus start by considering the following problem to classify only the local datasets $D^\ell_+$ and $D^\ell_-$.
\begin{prob} \label{prob:class_local}
	Find a function $f^{\ell}\in \mathcal{H}_k$ with the Gaussian kernel $k$ in (\ref{eqn:hatanaka_edit_gauss}) and $\gamma \in \mathbb{R}$ such that 
	\begin{subequations}
	\label{eqn:hatanaka_edit3}
	\begin{align}
		&D_{+}^{\ell} \subset \left\{ x \in \mathcal{X} \mid  f^{\ell} (x) + \gamma^{\ell} > 0  \right\}, \\
		&D_{-}^{\ell} \subset \left\{ x \in \mathcal{X} \mid  f^{\ell}(x)+ \gamma^{\ell} < 0  \right\}.
	\end{align}
	\end{subequations}
\end{prob}

Due to the property of RKHS, we have
\begin{align}
    f^{\ell}(d_i^{\ell}) = \langle f^{\ell}, k_{d_i^{\ell}}\rangle_{\mathcal{H}_k}
    \label{eqn:hatanaka_edit1}
\end{align}
for any data $d_i^{\ell} \in D^\ell$, where
$k_{d^{\ell}_i}(\cdot) = k(\cdot,d^{\ell}_i)$ and
$\langle \cdot, \cdot \rangle_{\mathcal{H}_k}$ is an inner product on the Hilbert space $\mathcal{H}_k$.
Since any function $f^c \in \mathcal{H}_k$ in the orthogonal complement of the space spanned by $k_d\ (d = d^{\ell}_1,\dots, d^{\ell}_{n^\ell})$ satisfies
$\langle f^c, k_{d_i^{\ell}}\rangle_{\mathcal{H}_k} = 0$
for all $d_i^{\ell} \in D^\ell$, we can assume
\begin{align}
    f^{\ell}(\cdot) = \sum_{j=1}^{n^\ell}c^{\ell}_j k_{d_j^{\ell}}(\cdot).
    \label{eqn:hatanaka_edit4}
\end{align}
The equation (\ref{eqn:hatanaka_edit1}) is then further rewritten as
\begin{align}
    f^{\ell}(d^{\ell}_i) = \sum_{j=1}^{n^\ell}c^{\ell}_j \langle k_{d_j^{\ell}}, k_{d_i^{\ell}}\rangle_{\mathcal{H}_k}
    = \sum_{j=1}^{n^\ell}c^{\ell}_j k(d^{\ell}_i,d^{\ell}_j).
    \label{eqn:hatanaka_edit2}
\end{align}
Accordingly,  (\ref{eqn:hatanaka_edit3}) are reformulated as the following linear constraints in $c^{\ell}_j\ (j=1,2,\dots,n^{\ell})$ and $\gamma^{\ell}$.
\begin{align}
	\label{eq:class_con}
	-\theta_{i}^{\ell} \left ( \sum_{j = 1}^{n^{\ell}} c^{\ell}_{j} k(d_{i}^{\ell}, d_{j}^{\ell}) + \gamma^{\ell} \right ) \leq -1 \ \ \forall i = 1,2,\dots, n^{\ell}. 
\end{align}

For organizing the problem in a vector format, define the matrix $K^{\ell} \in \mathbb{R}^{n^{\ell} \times n^{\ell}}$, whose $(i, j)$-element is $k(d_{i}^{\ell}, d_{j}^{\ell})$,  label matrix $\Theta^{\ell}:=\mathrm{diag}(\theta_{1}^{\ell}, \theta_{2}^{\ell}, \ldots, \theta_{n^{\ell}}^{\ell}) \in \mathbb{R}^{n^{\ell} \times n^{\ell}}$, and $c^\ell = (c^\ell_1,\dots,c^\ell_{n^\ell})$. Using these notations, (\ref{eq:class_con}) can be organized as $-\Theta^{\ell} [K^{\ell} \bm{1}_{n^{\ell}}] [(c^{\ell})^{\top} \gamma^{\ell}]^\top \leq - \bm{1}_{n^{\ell}}$, where $\bm{1}_{n^{\ell}} \in \mathbb{R}^{n^{\ell}}$ means the $n^{\ell}$ dimensional all-one vector.

Now, the local shape learning problem in Problem \ref{prob:class_local} is reduced to the following QP (\ref{eq:local_opt}), based on the concept of maximizing the margin of a hyperplane separating the dataset $D_{-}^{\ell}$ and $D_{+}^{\ell}$ in the RKHS [\cite{Seto:22}, \cite{Burges:98}].

\begin{align}
    \begin{split}
        \label{eq:local_opt}
    	\begin{array}{ccl}
    		\underset{c^{\ell} \in \mathbb{R}^{n^{\ell}}, \gamma^{\ell} \in \mathbb{R}}
    		{\mathrm{min}} && \frac{1}{2} (c^{\ell})^{\top} K^{\ell} c^{\ell} 
    		\\
    		\mathrm{subject~to} && - \Theta^{\ell} \begin{bmatrix} K & \bm{1}_{n^{\ell}} \end{bmatrix} \begin{bmatrix}c^{\ell} \\ \gamma^{\ell} \end{bmatrix}\leq - {\bm{1}}_{n^{\ell}}
    	\end{array}
    \end{split}
\end{align}

Following the target scenario, the function learned by each robot is expected to meet
\begin{align}
    \label{eq:f_ell_consensus}
    f^{\ell} = f\ \ \ \forall \ell = 1, 2, \ldots, N.
\end{align}
In the case of the polynomial kernel, the constraint is formulated by a finite number of equality constraints \citep{Oshima:22}.
On the other hand, the Gaussian kernel prohibits characterizing of each $f^{\ell}$ by a finite number of coefficients, and the approach of \citep{Oshima:22} is not applied.

\section{Distributed Shape Learning Using Gaussian Kernel}

In this paper, we assume that the robots share predefined grid points $g_1, g_2, \dots, g_M \in \mathcal{X}$.
We then fix the structure of $f^\ell$, instead of (\ref{eqn:hatanaka_edit4}), as
\begin{align}
    f^{\ell}(\cdot) = \sum_{j=1}^{M}c^{\ell}_j k_{g_j}(\cdot).
    \label{eqn:hatanaka_edit5}
\end{align}
Following the same procedure as (\ref{eqn:hatanaka_edit2}),
the equation (\ref{eq:class_con}) is changed to
\begin{align}
	-\theta_{i}^{\ell} \left ( \sum_{j = 1}^{M} c^{\ell}_{j} k(d_{i}^{\ell}, g_j) + \gamma^{\ell} \right ) \leq -1 \ \ \forall i = 1,2,\dots, n^{\ell}. 
	\label{eqn:hatanaka_edit6}
\end{align}

Since the space spanned by $k_d\ (d = d^{\ell}_1,\dots, d^{\ell}_{n^\ell})$ may not always be spanned by $k_g\ (g=g_1,g_2,\dots, g_M)$, the optimal solution to (\ref{eq:local_opt}) may not be represented in the form of 
(\ref{eqn:hatanaka_edit5}). Nonetheless, the function (\ref{eqn:hatanaka_edit5}) and $\gamma^{\ell}$ is a solution to Problem 2 as long as $c^{\ell}_j\ (j=1,2,\dots,M)$ and $\gamma^{\ell}$ are selected to meet (\ref{eqn:hatanaka_edit6}).
Despite the possible loss of optimality, (\ref{eqn:hatanaka_edit5}) is advantageous in sharing the function among robots.
Specifically, the constraint (\ref{eq:f_ell_consensus}) is simply described by a finite number of equality constraints
\begin{align}
    \label{eq:grid_points_consensus}
    \begin{bmatrix} c^{\ell} \\ \gamma^{\ell} \end{bmatrix} = z\ \ \forall \ell = 1, 2, \ldots, N,
\end{align}
where $c^{\ell}$ is redefined as $c^{\ell} = (c^\ell_1,\dots,c^\ell_M) \in \mathbb{R}^{M}$ and $\gamma^{\ell} \in \mathbb{R}$.
Hence, Problem \ref{prob:class} is formulated as the following distributed optimization problem under the above relaxation (\ref{eqn:hatanaka_edit5}). 
\begin{subequations}
\label{eq:dist_grid_points_opt}
\begin{align}
    		&\min_{z \in \mathbb{R}^{M + 1}, (c^{\ell}, \gamma^{\ell})_{\ell = 1}^N}
    		\sum_{\ell = 1}^{N}\frac{1}{2} (c^{\ell})^{\top} K^{\ell} c^{\ell} 
    		\\
    		&\mathrm{subject~to }\ - \Theta^{\ell} \begin{bmatrix} K & \bm{1}_{n^{\ell}} \end{bmatrix} \begin{bmatrix}c^{\ell} \\ 
    	\gamma^{\ell} \end{bmatrix}\leq - {\bm{1}}_{n^{\ell}}\ \  \forall \ell = 1,2,\dots, N\\
    		&\hspace{1.8cm}\begin{bmatrix} c^{\ell} \\ \gamma^{\ell} \end{bmatrix} = z\ \ 	\forall \ell = 1,2,\dots, N
    		\label{eq:dist_grid_points_opt_c}
\end{align}
\end{subequations}

The problem (\ref{eq:dist_grid_points_opt}) is easily reduced to so-called ADMM form. 
Introducing of new decision variables $x^{\ell} := (K^\ell)^{\frac{1}{2}} c^\ell$, $\beta_x \in \mathbb{R}_{>0}$, and a slack variable $y^\ell \in \mathbb{R}^{n^\ell}$, (\ref{eq:dist_grid_points_opt}) is transformed into
\begin{align}
    \label{eq:reformed_dist_grid_points_opt}
    \begin{split}
    		&\min_{z \in \mathbb{R}^{M + 1}, (x^{\ell}, \gamma^{\ell}, y^\ell)_{\ell = 1}^N}
    		\sum_{\ell = 1}^{N}\frac{\beta_x}{2} (x^{\ell})^{\top} x^{\ell}
    		\\
    		&\mathrm{subject~to }\ A^\ell \begin{bmatrix}x^{\ell} \\
    	\gamma^{\ell} \\ y^{\ell} \\ z \end{bmatrix} = \begin{bmatrix} - {\bm{1}}_{n^{\ell}} \\ 0 \end{bmatrix} \ \  \forall \ell = 1,2,\dots, N,\\
    		&\hspace{1.8cm} -y^\ell \leq 0\ \ 	\forall \ell = 1,2,\dots, N,
    \end{split}
\end{align}
where
\begin{align*}
    A^\ell := \begin{bmatrix}
        -\Theta^\ell \begin{bmatrix} K (K^{\ell})^{-\frac{1}{2}} & \bm{1}_{n^\ell} \end{bmatrix} & I_{n^\ell} && 0 \\
        - \begin{bmatrix}
           (K^{\ell})^{-\frac{1}{2}} & 0 \\ 0 & 1
        \end{bmatrix} & 0 && I_{M + 1}
    \end{bmatrix}.
\end{align*}
Furthermore, by multiplying the scaling matrix $Q^\ell \in \mathbb{R}^{(M + n^\ell + 1) \times(M + n^\ell + 1)}$ to both sides of the equality constraint in (\ref{eq:reformed_dist_grid_points_opt}) from the left, the distributed classification can be expressed in the same form as the equation (12) in \citep{Oshima:22}, except for $\gamma^\ell$.
Accordingly, the reformulated distributed optimization problem (\ref{eq:reformed_dist_grid_points_opt}) can be solved by a data-independent, performance-guaranteed continuous-time ADMM with the parameters and transfer functions designed based on the procedure presented in \citep{Oshima:22}.

Once the optimal solution $z_*,\ c^\ell_* = (c^\ell_{1*},\dots,c^\ell_{M*}),\ \gamma^\ell_*\ (\ell=1,2,\dots, N)$ of the original problem (\ref{eq:dist_grid_points_opt}) is obtained, each robot can identify the object shape as the curve 
\begin{align}
    \label{eq:learned_shape_grid_points}
    \begin{split}
        \left\{x\in\mathcal{X}\left|\ \sum_{j=1}^{M} c^{\ell}_{j*} k(x, g_j) + \gamma^{\ell}_{*} = 0 \right.\right\},
    \end{split}
\end{align}
which is common among all robots owing to the constraint (\ref{eq:dist_grid_points_opt_c}).

\section{Simulation Results}

In this section, we demonstrate the above solution through numerical simulation.

\begin{figure}
    \begin{center}
    \includegraphics[width=8.4cm]{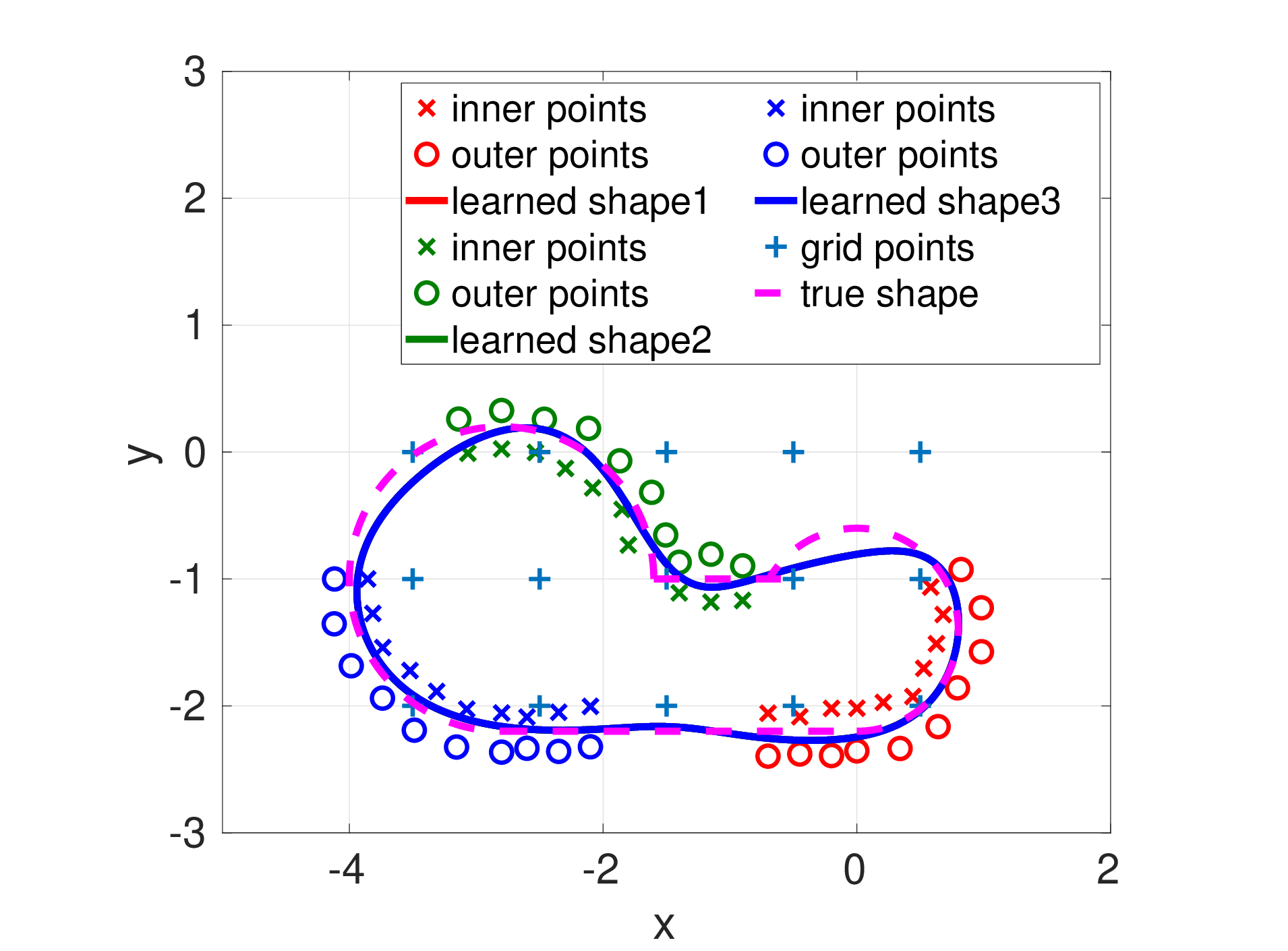}    
    \caption{The close curve with a dashed line is the target object.}
    \label{fig:learned_shape_5_3}
    \end{center}
\end{figure}

In this simulation, we consider three robots with $n^1 = n^2 = n^3 = 20$ data.
The object shape is illustrated by the magenta dashed curve, the data in $D_+$ and $D_-$ are by $\circ$ and $\times$ respectively, and the grid points are by the mark $+$ in Fig. \ref{fig:learned_shape_5_3}.
The color of the data corresponds to each robot.

\begin{figure}
    \begin{center}
    \includegraphics[width=8.4cm]{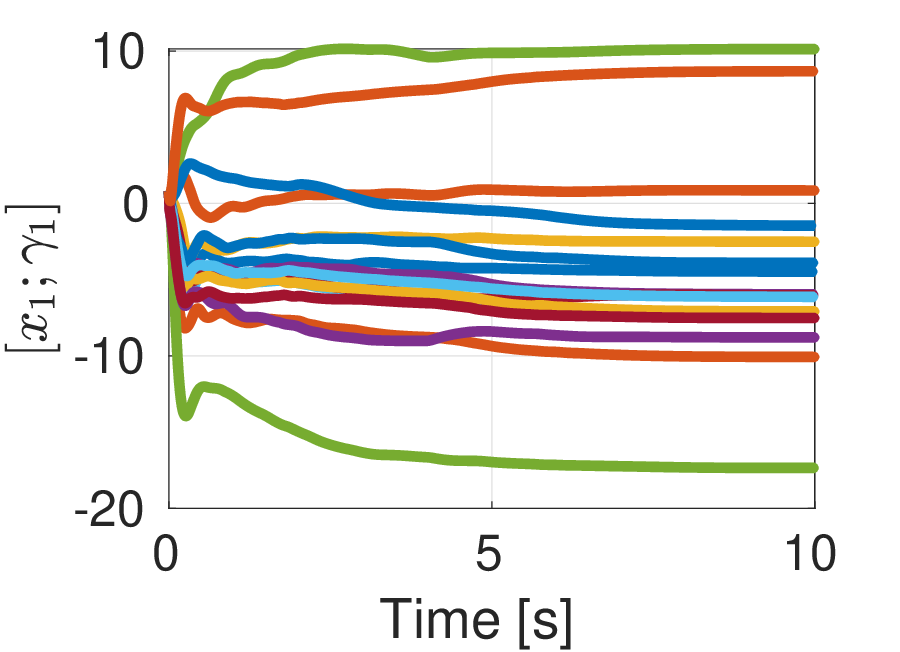}    
    \caption{Trajectory of $[(x^{\ell})^\top \ \gamma^\ell]$ for robot $\ell=1$. \label{fig:trajectory_x}}
    \end{center}
\end{figure}

Let us now apply the present solution to the above problem, where we set $T_{sp} = 0.01$s, sampling period $0.001$s and stop time $10$s, and other parameters are same as the examples shown in Table 2 of \citep{Oshima:22}.
The shapes learned by the robots are illustrated by the solid curves.
It is confirmed that all curves overlap, and the curves successfully separate $D_+$ and $D_-$.
Moreover, the learned shapes are similar to the target object.
The trajectories of the decision variables for robot $\ell=1$ are shown in Fig. \ref{fig:trajectory_x}, and the convergence time is around 500 $T_{sp}$.
They smoothly approach the optimal solution without any oscillation, owing to the loop shaping technique in (\cite{Oshima:22}).

\begin{figure}
    \begin{center}
    \includegraphics[width=8.4cm]{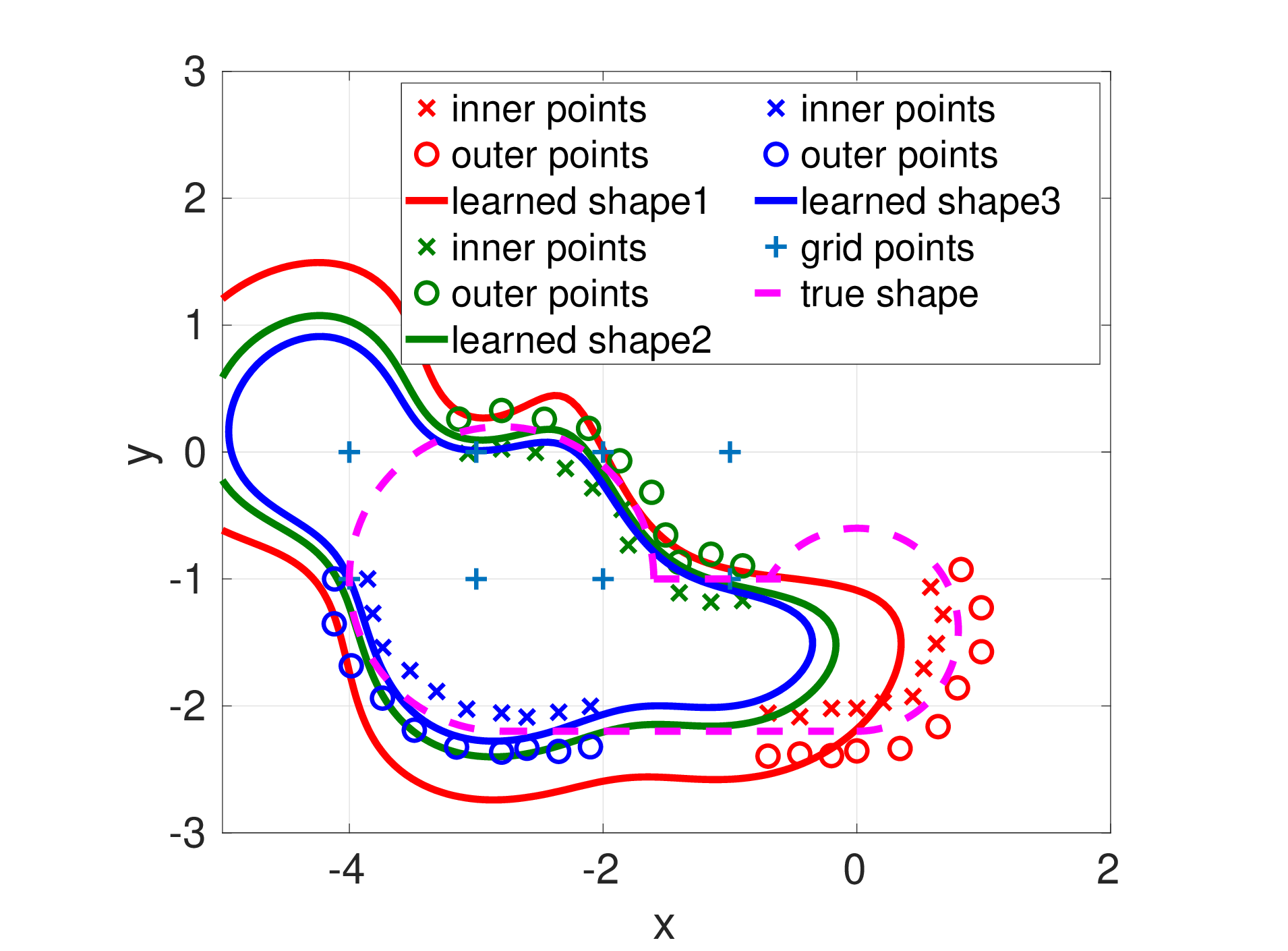}    
    \caption{The number and positions of grid points have been changed from Fig. \ref{fig:learned_shape_5_3}} 
    \label{fig:learned_shape_4_2_shifted}
    \end{center}
\end{figure}

Despite the ideal results above, the present solution still leaves open the issue of how to select the grid points.
Depending on the selection, the optimization problem may be infeasible and then the curves converge to the ones not separating the data as shown in \ref{fig:learned_shape_4_2_shifted}.
Appropriately choosing or adaptively changing 
the grid points are left as a future work.

\section{Conclusion}
In this paper, we addressed distributed learning of a complex object based on distributed optimization and kernel-based support vector machine with the Gaussian kernel.
In order that robots can share the object shape by a finite number of equality constraints, we reformulated the optimization problem assuming that the target function space is identified with the space spanned by the bases associated with not the data but a finite number of grid points.
The present approach was demonstrated through numerical simulations.

\bibliography{oshima}

\end{document}